[1]abiswas5@utk.edu, [2]liuy3@ornl.gov

# HUMAN–AI COLLABORATIVE AUTONOMOUS EXPERIMENTATION WITH PROXY MODELING FOR COMPARATIVE OBSERVATION

**Arpan Biswas[1]**
University of Tennessee-Oak Ridge Innovation Institute, University of Tennessee, Knoxville, TN 37996, USA

**Hiroshi Funakubo**
Department of Material Science and Engineering, Institute of Science Tokyo, Yokohama, 226-8502, Japan

**Yongtao Liu[2]**
Center for Nanophase Materials Sciences, Oak Ridge National Laboratory
Oak Ridge, TN 37830, USA

## ABSTRACT

*Optimization for different tasks like material characterization, synthesis, and functional properties for desired applications over multi-dimensional control parameters need a rapid strategic search through active learning such as Bayesian optimization (BO). However, such high-dimensional experimental physical descriptors are complex and noisy, from which realization of a low-dimensional mathematical scalar metrics or objective functions can be erroneous. Moreover, in traditional purely data-driven autonomous exploration, such objective functions often ignore the subtle variation and key features of the physical descriptors, thereby can fail to discover unknown phenomenon of the material systems. To address this, here we present a proxy-modelled Bayesian optimization (px-BO) via on-the-fly teaming between human and AI agents. Over the loop of BO, instead of defining a mathematical objective function directly from the experimental data, we introduce a voting system on the fly where the new BO samples (high-dimensional experimental descriptors) will be compared with existing samples, and the human agents will choose the preferred samples. These human-guided comparisons are then transformed into a proxy-based objective function via fitting Bradley–Terry (BT) model. Then, to minimize human interaction, this iteratively trained proxy model also acts as an AI agent for future surrogate human votes. Finally, these surrogate votes are periodically validated by human agents, and the corrections are then learned by the proxy model (adjusting function space and AI agent) on-the-fly. We demonstrated the performance of the proposed (px-BO) framework into simulated and BEPS data generated from PTO sample. We find that our approach provided better control of the domain experts for an improved search over traditional data-driven exploration. Over (px-BO) iterations, we see that while proxy-modelled AI agents accelerate exploration via providing highly accurate decisions to compare learned patterns of physical descriptors, human agents help to steer the AI agents in learning any new preferable pattern of physical descriptors. Overall, our method signifies the importance of human-AI teaming in an accelerated and meaningful material space exploration.*

## 1. INTRODUCTION

Autonomous experimentation and/or self-driving laboratories have increasingly advanced the throughput and efficiency of materials research by integrating robotics and artificial intelligence (AI)/machine learning (ML). The core of autonomous experiments (AEs) is a closed optimization or discovery loop, starting from performing experiment, measuring outcome, to parameter-outcome relationship analysis by AI/ML and decision-making algorithms to select next experiment. One of the most popular approaches in the design of AE is Bayesian optimization (BO). Bayesian Optimization (BO)[1], [2], [3] is an active learning method which aims to autonomously explore the parameter space and continually learns the unknown ground truth and its global optimal region, where the ground truth function can be either black-box or expensive to evaluate, or both. Given a few evaluated training samples, the expensive unknown function is replaced by a cheaper surrogate model (*e.g.* Gaussian Process)[4], [5], [6], and the surrogate model continues to learn the human defined region of interests with adaptive selection (*e.g.* Acquisition function)[7], [8], [9], [10] of locations for future expensive evaluations. BO is more popular than other designs of experiment methods (*e.g.* Random sampling, Latin Hypercube sampling, *etc.*) as it is designed to converge to the optimality with minimal expensive evaluations. Thus, BO has attracted special attention in the materials science domain where accelerated BO driven autonomous discovery has been particularly impactful, enabling efficient identification of optimal conditions for particular material properties without human intervention, such as bandgap optimization[11], small-molecule emitters discovery, maximizing carbon nanotube growth rates[12], and so on. This type of autonomous workflow with BO has been widely used in recent studies to adaptively explore expensive control parameter spaces of physical/simulation models[13], [14], [15], [16], [17], [18] and experiments[19], material structure-property relationship discovery[20] and to develop autonomous platforms towards accelerating chemical[21], [22] and material design[23], [24], [25]. Recently, interactive BO frameworks through minor human intervention (human in the loop) proved to have better material processing[26] and microscope experimental steering[27], [28], [29], [30], [31], [32]. A number of excellent reviews on BO are available[1], [33], and it is now implemented in a broad range of Python libraries including BOtorch (Pytorch)[34] and Gpax (Jax)[35]. In most current AE systems, the parameter-outcome relationship analyses rely on scalar metrics of outcomes such as peak intensity, band gap, or conductivity. These metrics often are not direct experimental observations, instead they are compressed numerical values extracted from raw data such as spectra, microscopy images, or multidimensional measurements. As such, AE systems are driven by what can be numerically evaluated, and the ML algorithm does not learn from the experiment directly, it learns from the scalar metric that defines the objective of an AE.

This introduces a fundamental limitation. The scalar metric requires a predefined data-reduction procedure that is implemented as analysis scripts prepared before AEs begin. This procedure is based on the observation in the AE is known in advance and that it can be explicitly captured by a scalar metric. However, AE systems often explore large unknown materials spaces, such as in unfamiliar regimes, AE outcomes very likely either violate the assumptions underlying the predefined metric or the predefined metric cannot represent the experimental observation. This introduces artifacts and/or limitations in ML algorithm learning parameter-outcome relationship. In this sense, AE systems are not limited by insufficient data but by misrepresentation of information.

In addition, many scientifically valuable outcomes cannot be represented by scalar metrics at the discovery stage. At early stage of discovery, researchers identify promising directions by recognizing unusual patterns or forming hypothesis via comparative evaluation. In our earlier work [27], [28], we developed Bayesian Optimized Active Recommender System (BOARS) that allows researchers to identify interesting patterns at the early exploration of AE. Unlike specifying a target objective function to optimize, BOARS provide us with designing the target functions via human on-the-fly voting during the process of BO/AE exploration. With enough human satisfactory early explorations, once the target function is confirmed, the human-guided exploration is switched to fully autonomous exploration to accelerate the learning of the parameter space. In previous work, we have also provided similar approach to design the constraint boundary (separating clean vs noisy experiments) at the early stage of exploration via on-the-fly human voting[36], [37]. However, the limitation of these developed human-AI tools is the restriction of human intervention to only early stages of exploration, assuming there are higher chances of identifying novel experiments in the early stages. In reality, such material spaces are vast to explore and therefore such novel experiments can be sighted in later stages as well, which can be ignored. On the other hand, it is also not feasible for human adjustment in every iteration of BO, as that will enormously increase the overall cost of AE. Thus, a better balance is required to integrate human knowledge into AE. Moreover, there is currently very limited systematic approach to translating human comparative judgments into a numerical representation that can be learned and utilized by ML models. Thus, valuable qualitative insights provided by expert evaluation cannot be effectively incorporated into AE decision-making process.

In this work, we develop an approach of a proxy-modelled Bayesian optimization (px-BO) via on-the-fly teaming between human and AI agents. Here, we design a utility function to convert human comparative judgements into numerical proxy, that can be directly used by prediction models in BO. Therefore, human experts' comparative and qualitative evaluation of experimental outcomes based on multiple data modalities, prior knowledge, and contextual reasoning are transformed into quantitative evaluation to integrate into the BO framework. Over the px-BO iterations, to improve the balance between exploration alignment and computational cost (via human intervention), this proxy model is also utilized as an AI agent to represent human qualitative evaluation of new experimental outcomes, which is periodically validated and adjusted. Thus, the goal of the proposed px-BO is to 1) integrate human qualitative assessment into BO for better exploration and 2) reducing the iterative human intervention to accelerate scientific discovery. To note, this utility function could be integrated with any type of data including images, spectra, or multimodal data; this utility function can also be integrated with other active learning algorithms available in AE field. We integrated this approach with both traditional Gaussian Process (GP)-based and Deep Kernel Learning (DKL)[38]-based BO and demonstrated the application in microscopy field, over simulated and experimental Band Excitation Piezoresponse Spectroscopy (BEPS) data, for analyzing the structure-property relationship.

The roadmap for this paper can be stated as follows: Section 2 provides the detailed description and algorithm of the px-BO architecture. Section 3 showcased the implementation of the prototype to simulated and experimental BEPS data of a ferroelectric thin film (i.e., $PbTiO_3$), with performance analysis between fully human to human-AI agent qualitative assessments. Section 4 concludes the paper with final thoughts.

## 2. METHODOLOGY

The px-BO architecture has four key developments – 1) the human operated voting system to compare between two experimental outcomes, 2) the transformation of qualitative human-guided comparative data into a quantitative utility, 3) design and continual training of an AI agent with human-guided comparative data to mimic human operated voting system and 4) periodic human-guided validation of the AI agent for quality control.

In the context of AE in material characterization, given a location in the material image space, the spectral structure is captured first and visualized. As stated, this spectral structure is enriched with several key material properties, which is considered for optimization to achieve the desired application. Then, on-the-fly, the human compares the current sampled spectral structure with few randomly chosen explored samples and the current best explored samples. With several comparisons between new and existing samples, we fit the data into a statistical model such as Bradley–Terry model (B-T)[39], [40]. B-T is a probability model for the outcome of pairwise comparisons between items, teams, or objects. B-T has been utilized successfully as reward modelling based on limited pairwise comparison data. Given a pair of items $i$ and $j$ drawn from some population, it estimates the probability that the pairwise comparison $i > j$ turns out true, which further estimates the utility of $i$. Higher score of $u(i)$ means the item $i$ is highly preferable. Though B-T models are popular in application to domains such as sports analytics, social science and marketing[41], where consumer preferences and ranking orders are common, it has very limited to no application in AE. In the case of AE, we can think of the items as the spectral structures at given locations in the material image space. It is to be noted that it is extremely costly to calculate the exact utility of a given sample as it needs to be compared with all other existing samples, thus the number of human voting increases with the number of px-BO iterations. To eliminate this significant increase in cost of voting with the progression of the px-BO explorations, we considered a fixed number of voting in each iteration by selecting the subset (randomly) from the existing samples. Thus, we calculate the estimated utility based on the subset of comparisons, for the new sample $x_{new}$. Also, all the estimated utilities of existing samples $X$ are updated based on the new comparisons.

Choosing the subset of samples for comparison, a new captured sample $x_{new}$ is always compared with the current best sample $x^+$ to look for any novel (not identified in previously explored locations) structures that the human prefers to investigate further (steering the exploration). Compared with other random samples $X_C$ opens the possibility that though the new samples are not preferable compared to the current best but still can be highly preferred to other existing samples, thus it might be worth exploiting. In scenarios where the new sample $x_{new}$ are rarely preferred over existing samples, the estimated utility $u(x_{new})$ will be lower. However, with new set of comparisons at future px-BO iterations, the new sample $x_{new}$ still could improve the utility via re-estimation. Similarly, the new sample $x_{new}$ score higher utility can get lower with future data augmentation. This is a key step to the design and continual learning in the AE as any earlier inaccuracies (either human choices or biased subset $X_C$) preferred or non-preferred experimental outcomes (when explored locations are less) can be corrected with future experiments.

Once the human voting system is designed for comparing different experimental outcomes, the next step is to improve the balance of the design with minimal human intervention. In other words, if a human needed to provide multiple votes in each iteration of px-BO, it increases the complexity of AE significantly and limits the application for accelerated discovery. Thus, we aim to integrate an AI agent to mimic the human voting system. In this case, we consider the trained BT model to act as an AI agent. Given a new captured sample $x_{new}$, we first select the proxy sample $x_{px} \subset X$ from the explored samples, based on highest structural similarity between $x_{new}$ and $x_{px}$. Then, knowing the utility of proxy sample $u_{px}$, the voting choices can be estimated between $x_{px}$ and $X_C$. These surrogate comparisons are then augmented with existing comparison data $D_C = [D_C; \bar{c}_{x_{new}}]$ and the utilities of all the explored samples $D = [X, x_{new}; U, u_{new}]$ are re-estimated. To design the quality control of the AI agent, we introduce periodic domain-expert validation of the surrogate votes. Here, the domain expert visualizes all the current non-validated surrogate votes together. Then, based on the assessment, the domain expert can specify over a single prompt which voting numbers need to be changed. The algorithm switches the preference of the votes from $(x_i, x_j)$ to $(x_j, x_i)$ where the first location is preferred over second location. Once all the corrections are made, $D_C$ is corrected and eventually $D$ is re-estimated. The integral architecture of BO remains the same and therefore we have not highlighted here but directly explained in the detailed algorithm (**Table 1**).

Previously, human-augmented AE have been developed in microscopy in accelerating meaningful discoveries in different field of applications such as rapid validation of thousands of biological objects or specimen tracking results [42], rapid material discovery of novel lithium ion conducting oxides through synthesis of unknown chemically relevant compositions (CRCs) [43], autonomous synthesis with pulsed laser deposition for remote epitaxy of BaTiO$_3$ growth[44], human-AI collaboration for accelerated workflows design and preparation[45], and human assessment of raw data for quality control in both autonomous synthesis and autonomous characterization[46], [47]. In summary the contributions of the paper are as follows:

a) We propose a computational proxy modelling for human-guided comparative observation in AE to improve AI alignment, by allowing the experimentalists to compare the quality of the captured spectra during microscope measurements, and then design a proxy model to estimate utility of the experiments and act as an AI agent to automate human voting process.
b) We connect the proxy model with GP-BO and DKL-BO workflows to guide the adaptive search process in the multi-parameter space for autonomous material structure-property relationship learning.
c) Implementation of the prototype on the simulated and experimental data samples of ferroelectric thin films as proof of concept and analyze the results.

**Figure 1** provides the overall design of the px-BO system. **Table 1** shows the detailed algorithm of the px-BO.

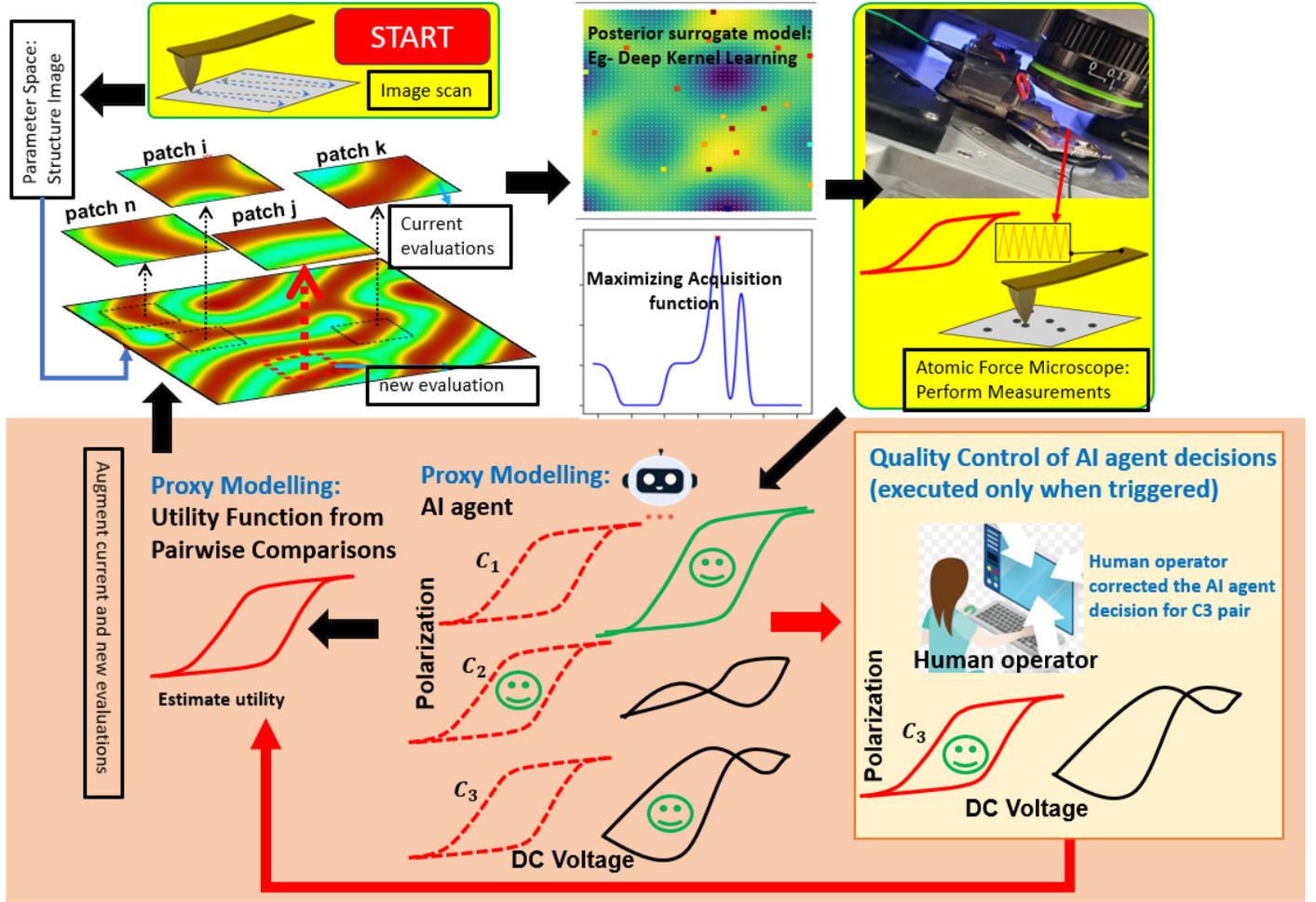

**Fig 1.** Proposed architecture of the proxy-modelled Bayesian Optimization (px-BO). The highlighted section in orange is the contribution in this paper. This figure is a reference to implementation in microscopy, however, can be designed for any problem where human guided qualitative comparisons are made in the loop. Here, starting with scanning a material sample in the microscopy, we aim for autonomous material characterization, where given some image patches autonomously collected, is fitted with the prediction model (eg. DKL), which in turn define the acquisition function (eg. EI). Then, maximizing this acquisition function, the next location of image patch is selected, and microscopy measurement is done. Then, the new measurement (eg. red spectral structure) is compared with a subset of the earlier measurements via proxy-modelled AI agent. The green spectral structure is the current best outcome whereas the black spectral structures are non-optimal earlier measurements. The dashed red spectral is the proxy spectral of the new red spectral (highest structural similarity). Then, if human validation is triggered (steps with red arrows), the human validates all the current non-validated AI agent proxy decisions, and re-correct if required. The new proxy (AI agent) or validated (human agent) new comparisons augments with previous comparison data, and the proxy-modelled utility values are calculated for the measured locations. This loop continues till convergence.

*Table 1: Algorithm: Proxy-modeled Bayesian Optimization (px-BO)*

1. **Segmentation of local image patches**:
   a. Choose a material sample. Scan topographic image on the microscope (eg Atomic Force Microscope (AFM)).
   b. For DKL-BO, segment the image into square patches of window size, $w$.

2. **Initialization for px-BO:** State maximum iteration, $M$. Set number of initial samples, $j$, total number of initial comparisons, $n \times j$, human-validation of surrogate votes after every $m \ll M$ iterations. Randomly select $j$ samples, $X$. *We highlight this step as the contribution in this paper.*
   a. For sample $i$ in $j$, get spectral data, $S_i$, measured from microscope.
   b. For each sample $i$ in $j$, user compares with randomly selected $n$ samples ($n < j - 1$). User prompts the sample number preferred between the two samples in each comparison. The user choices are stacked as comparison data, $D_c = [x_{i=1}, x_{c1}]; \ldots; [x_{i=j}, x_{cn}]$ where the first element $[x_{i=1}, x_{c1}]$ is the preferred choice. It is to be noted that the placement of the 1st and 2nd elements for each comparison can be different, and this is only provided as a reference.
   c. **Proxy Modelling for Comparative Data**: Given $D_c$, fit **Bradley-Terry B-T model** $\Upsilon$. Estimate utilities for $j$ samples in $X$, as $U(X) = \Upsilon(D_c)$.
   d. **Build Dataset**, $D_j = \{X, U\}$ with $X$ is a matrix with shape $(j, p)$ and $U$ is the array with shape $(j)$ where $p$ is the number of design parameters. For DKL-BO, $X$ is a matrix with shape $(j, w \times w)$.

**Start BO.** Set $k = 1$. For $k \leq M$

3. **Surrogate Modelling**: Fit a prediction model for BO. Here, we considered Gaussian Process (GPM) and Deep Kernel Learning (DKL) models. Optimize the hyper-parameters of kernel functions for each of the surrogate models.
4. **Posterior Predictions**: For each GP model, compute posterior means and variances of the utilities $\mu(U), \sigma^2(U)$ for the unexplored locations $X_{ue}$.
5. **Acquisition function:** Compute the acquisition function value $\psi$, given $\mu(U), \sigma^2(U)$. Here, we have considered *Expected Improvement (EI)* Acquisition function. The mathematical description of EI is given as below **Eqs. 1-2.** $u^+$ is the current best utility score, $\Phi(.)$ is the cdf; $\phi(.)$ is the pdf; $\xi \geq 0$ is a small value which is set as 0.01. Maximize acquisition function $\max_{X_{ue}} \psi$, to select next best location, $x_{new} \subset X_{ue}$ for evaluations.

$$\psi(x) = \begin{cases} \left(\mu(u(x)) - u^+ - \xi\right) \times \Phi(Z, 0, 1) \\ + \sigma(u(x)) \times \phi(Z) \; if \; \sigma^2(u) > 0 \\ 0 \; if \; \sigma^2(u) = 0 \end{cases} \quad (1)$$

$$Z = \begin{cases} \frac{\mu(u(x)) - u^+ - \xi}{\sigma(u)} \; if \; \sigma^2(u) > 0 \\ 0 \; if \; \sigma^2(u) = 0 \end{cases} \quad (2)$$

6. **New Measurements and Human-AI agent guided comparisons:**
*We highlight this step as the contribution in this paper.*
   a. At the new location $x_{new}$, get spectral data, $S_{new}$.
   b. **Generate a subset of explored locations to compare with new locations**: A subset $X_C \subset X$, containing $q$ locations, are selected where the 1st element is the current best location $x^+$ and the remaining $q - 1$ locations are randomly chosen.
   c. **Find the proxy for the new location:** Calculate the structural similarity, $\Delta$, between $x_{new}$ and all the locations in $X$. $\Delta(x_{new}, x)$ is computed from the function "structural_similarity" in "skimage.metrics" Python library. Maximize $\max_X \Delta$ to find the proxy location $x_{px}$.
   d. **Proxy Modelling for AI agent of Human Voting System:** Utilize the trained B-T model $\Upsilon$ (step 2c) to estimate the probabilities of the preference of the proxy location $x_{px}$ between each location in $X_C$. Build the surrogate comparison matrix $\bar{c}_{x_{new}} = [x_{px,k}, x_k^+]; [x_{px,k}, x_{1,k}]; ....; [x_{px,k}, x_{q-1,k}]$. It is to be noted that the placement of the 1st and 2nd elements for each surrogate comparison by the AI agent can be different, and this is only provided as a reference.
   e. **Periodic Human-Validation of AI agent decisions:** This step will only occur after every $m \ll M$ as set in Step 2. During the BO loop, when this step is triggered, a prompt is given to request domain expert to check all the current non-validated surrogate comparison matrix $\bar{c}_{x_{new}}$, pre-specified by the location serial number. For the surrogate votes that need to change, the domain expert will enter the location serial numbers in a single prompt. Then, for all those incorrectly assessed locations, the surrogate comparison will switch from $[x_{px,k}, x_k^+]$ to $[x_k^+, x_{px,k}]$. For all those accurately assessed locations, the surrogate comparison will be same as earlier such as $[x_{px,k}, x_k^+]$. After all the corrections, we have the validated human-surrogate teamed comparison matrix $c_{x_{new}}$. It is evident to say, to avoid the error of assessment from structural dissimilarity, the domain-expert assess via visualizing spectral $S_{new}$ of actual $x_{new}$ instead for the proxy $x_{px}$.

7. **Augmentation and Function Evaluations:** Augment Comparison data as $D_{c,j+k} = [D_{c,j+k-1}; \bar{c}_{x_{new}}]$ if step 6e is not triggered or $D_{c,j+k} = [D_{c,j+k-1}; c_{x_{new}}]$ if step 6e is triggered. Retrain the proxy B-T model $\Upsilon$ with $D_{c,j+k}$. Augment $x_{new}$ in $X$. Re-calculate utilities for all samples in $X$ as $U(X_{j+k}) = \Upsilon(D_{c,j+k})$. Repeat step 3-7 till convergence.

## 3. RESULTS

Here, we showcase the demonstration of the proposed px-BO architecture for autonomous exploration over 1) simulated and 2) BEPS microscopy dataset of ferroelectric PTO thin film samples.

**Fig. 2** shows the generated images with numerically driven quantitative output to measure the preference of the images. The images were generated by a variational autoencoder, which is trained by a set of PbTiO3 PFM phase image patches with various domain structures. In the generation of this image set, we varied the latent variable along one dimension that is related to domain wall location and curvature. This gives us a set of images allows human to do qualitative evaluation, while the latent variable used for generation can also serve as a quantitative metric to validate the performance of px-BO. In the series of images (Fig. 2a.), we derive the numerical scores (Fig. 2b.) on the preference of the images in terms of physical knowledge. These physical knowledge-based preferences are- a) the observed image should contain only one domain wall, b) the domain wall should be located near the center of the image and c) the domain wall should be relatively straight overall. For autonomous exploration, we choose 10 initial random samples with 20 initial random comparisons. We used Gaussian Process for the BO prediction model. For Step 6b in Table 1, we choose $q = 3$. We ran the model for 20 BO iterations.

**Fig. 3** shows the performance of the px-BO model with fully human voting (each BO iteration). Fig. 3a are some of the initial voting made by human between the paired images (in columns), with the preferred images as indicated with the smiley figure (as per the stated physics knowledge). From the initial 20 comparisons over 10 images, the image associated with the double smiley obtain the highest initial utility. Figs. 3b-d, are the votes made on-the-fly during the px-BO exploration, where the right of each paired image is the new selected image and the left images are the explored images. From fig. 3c and d, we could discover a better image than the current best images, as the autonomous exploration progress. Figs. e and f show the estimation of the GP after the maximum number of iterations are conducted. As expected, we can see exploitation in the high-preference image locations and being mostly aligned in the good and bad regions with the numerical ground truth. However, interestingly, if we closely observe the optimal region (region surrounding the highest peak in fig e and g), there is a slight drift in the optimal location between the numerically driven and proxy-modelled estimated output. As we compare the images at those two optimal locations, as suggested numerically (location 1) and proxy-modelled (location 2), we can clearly see the proxy-modelled based optimal image is physically superior with less noisy domain walls than the numerically driven optimal solution.

**Fig. 4** shows the performance of the px-BO model with human-AI agent collaborative voting. In this case, human validation is triggered after every 4 BO iterations, i.e. after every 12 AI-agent votes. In this case, we started with same initial assessment as shown in Fig. 3a. Figs. 4a-f, are some of the votes made on-the-fly during the px-BO exploration, where the right of each paired image is the newly selected image and the left images are the explored images. From fig. 4a to c, the AI agent provides accurate assessment, as indicated with the dashed smiley. However, Fig. 4d, we can see an inaccurate assessment made, which is afterwards corrected with human validation (see fig. 4e). With corrections and continual learning, the AI agent gets better and is even able to discover a physically better image, as compared with the current best image (see fig. 4f). Figs. 4g and h show the estimation of the GP after the maximum number of iterations are conducted. Here also, we can see exploitation in the high-preference image locations, being mostly aligned in the good and bad regions with the numerical ground truth, and a slight drift in the optimal location between the numerically driven and proxy-modelled estimated output. As we compare the images at those two optimal locations, as suggested numerically (location 1) and proxy-modelled (location 2) (as in fig. 4i), we can clearly see the proxy-modelled based optimal image is physically superior with less noisy domain walls than the numerically driven optimal solution. Comparing the optimal solution with human-AI teaming in fig 4 with the optimal solution in fully human voting in fig 3, we see similar solution achieved with AI agent proxy voting, while reducing the intervention of human significantly.

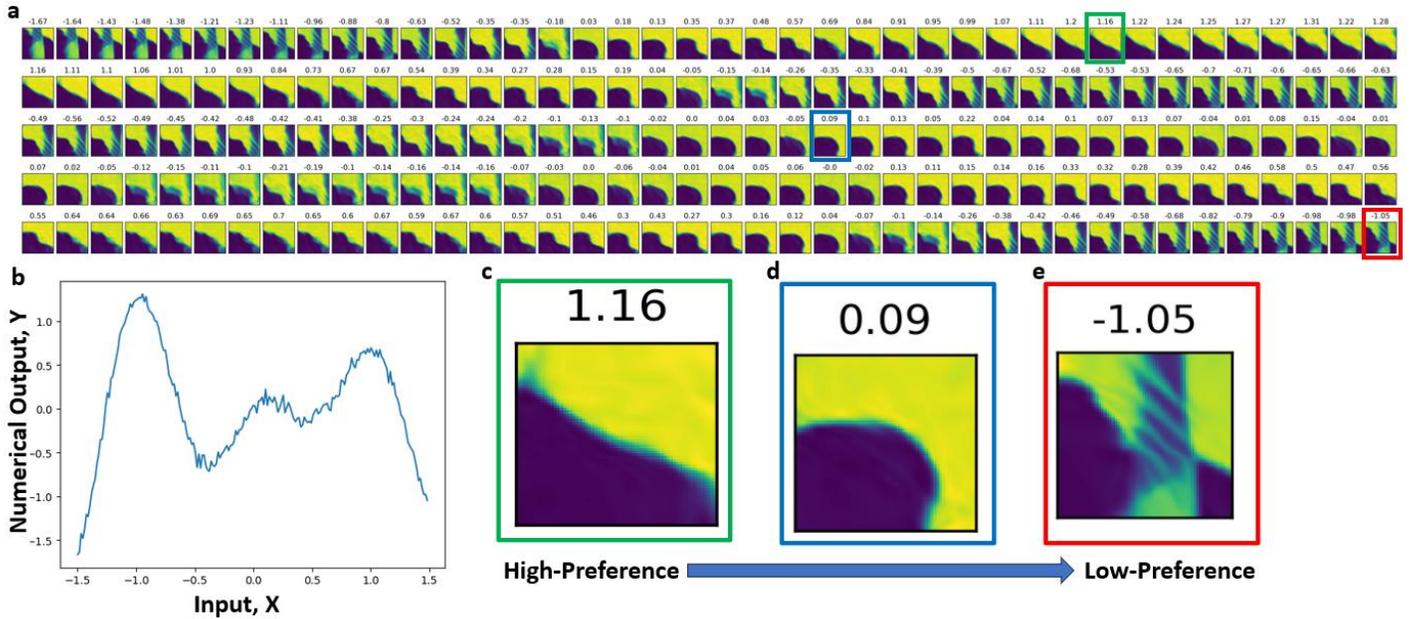

**Fig. 2.** Simulated data of ferroelectric material generated in Variational Autoencoder. Based on the series of images in (a), the physics driven image quality score is defined as 1) the observed image should contain only one domain wall, 2) the domain wall should be located near the center of the image and 3) the domain wall should be relatively straight overall. Fig. (b) is the numerical output based on these quality scores.

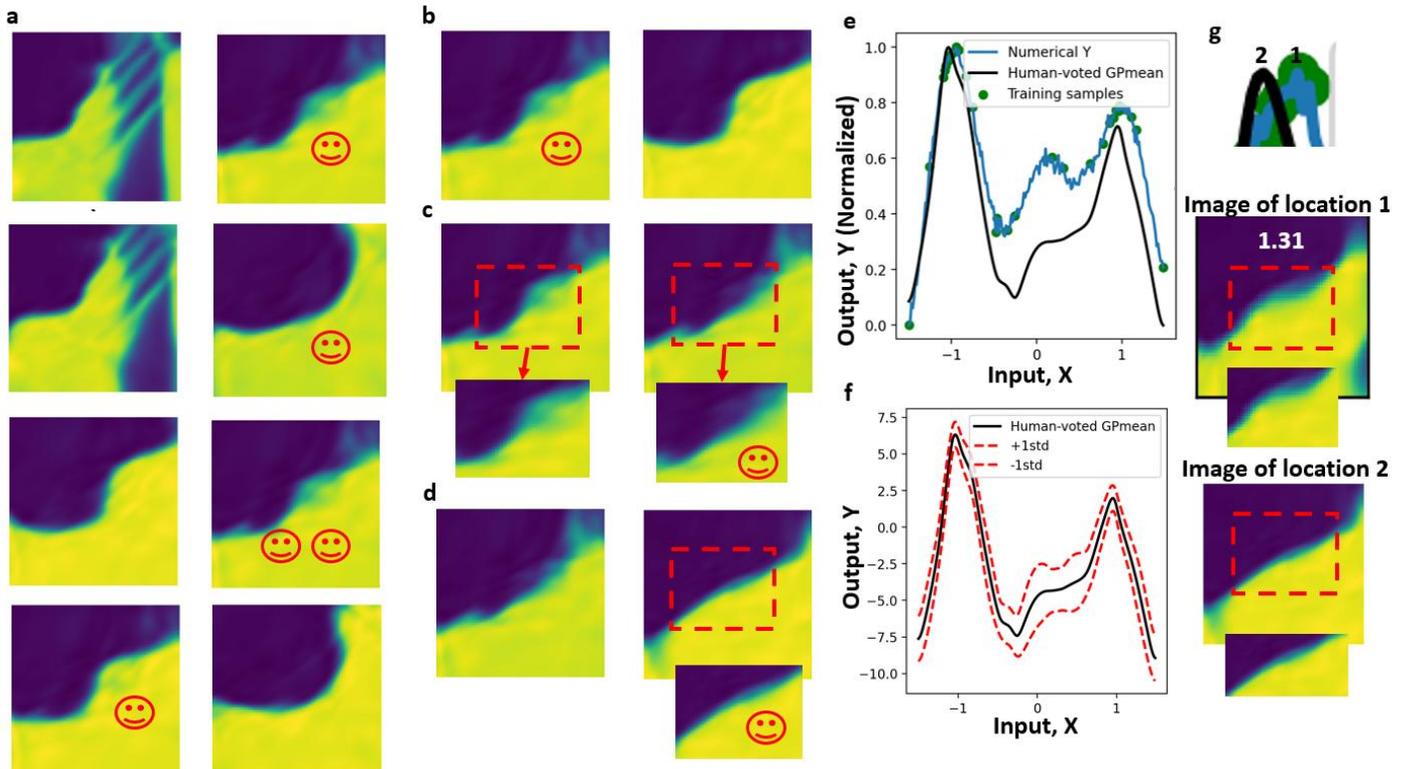

**Fig. 3.** Simulated ferroelectric data: Performance of the px-BO architecture where proxy modeled is conducted from comparative observation, based on fully human assessment. Here, the proxy model is not used to design the AI agent. Fig (a) are the series of initial paired comparison, as human prefer images indicated by the associated smiley. Figs (b) –(d) are the manual comparison during the px-BO exploration. Figs (e)-(g) are the final prediction and the performance comparison of the px-BO.

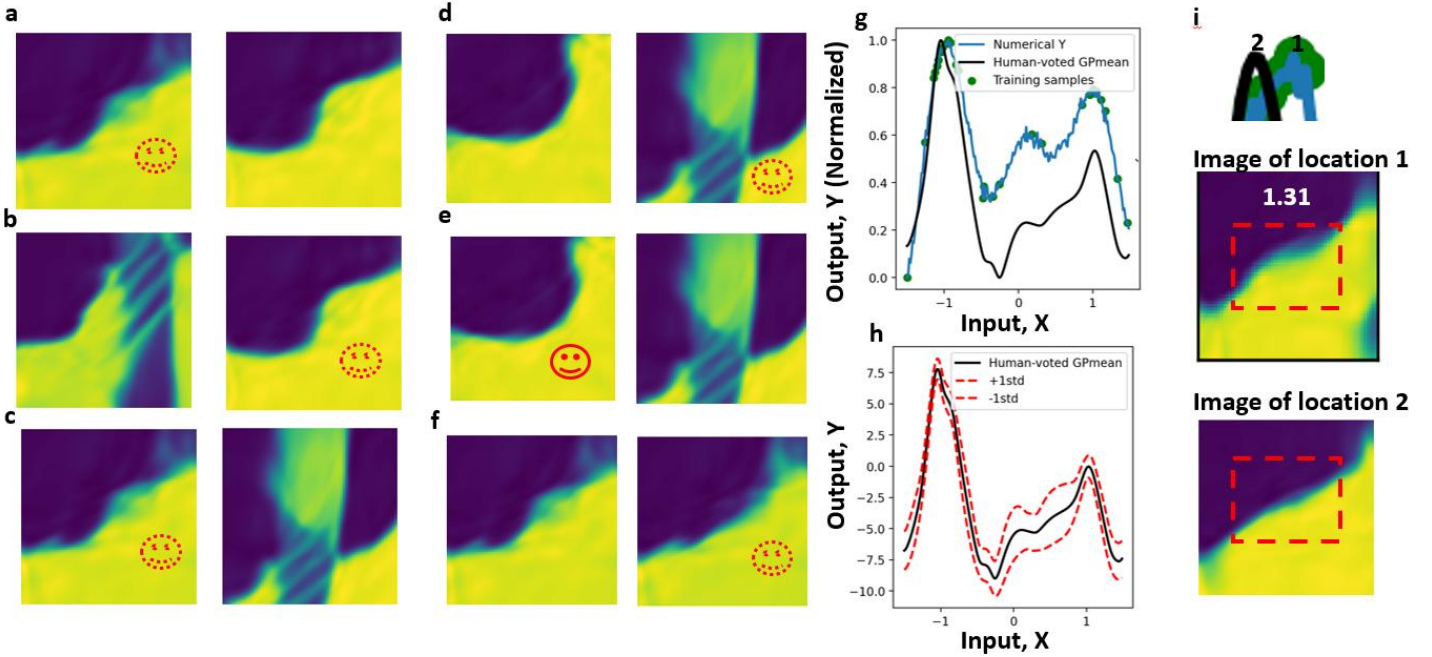

**Fig. 4.** Simulated ferroelectric data: Performance of the px-BO architecture where proxy modeled is conducted from comparative observation, based on human-AI teaming. Here, the proxy model is used to design the AI agent. The initial data and comparison are similar to previous analysis as in fig. 3a. Figs (a) –(f) are the comparison during the px-BO exploration, as done by AI agent associated with dashed smiley and human as associated with solid smiley. Figs (g)-(i) are the final prediction and the performance comparison of the px-BO.

The second set of analysis we did over experimental BEPS data of PTO thin film sample, which consists of image data and spectroscopy data, allowing researchers to explore the structure (from image) and property (from spectroscopy) analysis. Here, we aim to find spectral structure with higher loop area and lower experimental noise or variation among each loop. For autonomous exploration, we choose 10 initial random samples with 20 initial random comparisons. We used deep kernel learning (DKL) for the BO prediction model. For Step 6b in Table 1, we choose $q = 3$. We ran the model for 50 BO iterations. **Figure 5 and 6** shows the results of the px-BO exploration over two structure (design) spaces of PTO thin film sample, with comparison among quantitative loop area computation vs qualitative comparison observation without and with AI-agent teaming.

From fig. 5a, the left figure is the structure image exploration space, where at each co-ordinate (pixel point), a spectral structure can be characterized. From the spectral structure, we can calculate the numerical loop area which is mapped in the right figure of 5a. Fig. 5b is one of initial comparisons done by human where the top figure is the most preferred from initial comparison. Figs 5c and d show the final prediction and uncertainty map of the px-BO after 50 iterations and 150 fully manual and human-AI collaborative comparisons (human validation is triggered after every 15 AI-agent votes) respectively. In both cases, we can see the location of the optimal solutions (indicated by red arrows) are in similar locations, but different to the locations (indicated by black arrows) as suggested from numerical calculations. Comparing the spectral structure at those locations indicated by black and red arrows, we can clearly see that the proxy-modelled based optimal spectra are significantly physically superior with less noisy loops than the numerically driven optimal spectra.

From fig. 6a, the left figure is the structure image exploration space, where at each co-ordinate (pixel point), a spectral structure can be characterized. Fig. 6b is one of initial comparisons done by human where the top figure is the most preferred from initial comparison. Figs 6c and d show the final prediction and uncertainty map of the px-BO after 50 iterations and 150 fully manual and human-AI collaborative comparisons (human validation is triggered after every 15 AI-agent votes) respectively. In both cases, we can see the location of the numerically driven optimal solutions (indicated by 1) and qualitative comparison driven optimal solutions (indicated by 2) are in similar locations. However, comparing the spectral structure at those locations, we can clearly see that the proxy-modelled based optimal spectra are slightly physically superior with less noisy loops than the numerically driven optimal spectra.

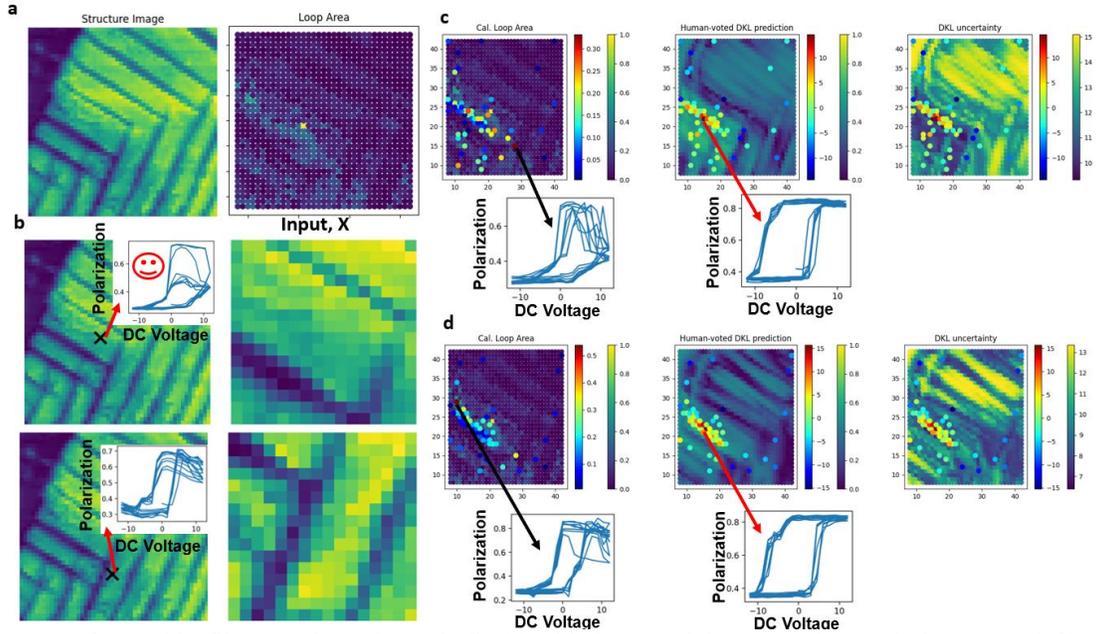

**Fig. 5.** BEPS dataset 1 of PTO thin-film sample as shown in fig (a) with the material structure image design space (left) and the relatively numerically calculated material property such as loop area (right). Fig (b) shows a sample initial comparative human assessment with the preferred location (image patch in the right images) associated with smiley. Figs (c) and (d) are the performance of the px-BO architecture with fully manual and human-AI agent teaming assessments. The left images of these figures are the numerically calculated loop area with the dots representing the same loop area of the explored samples. The middle and the right images of these figures are the predicted mean and variance maps of the utility based on the qualitative assessments (focused on clean spectral with high loop area) with the dots representing the utility of the explored samples.

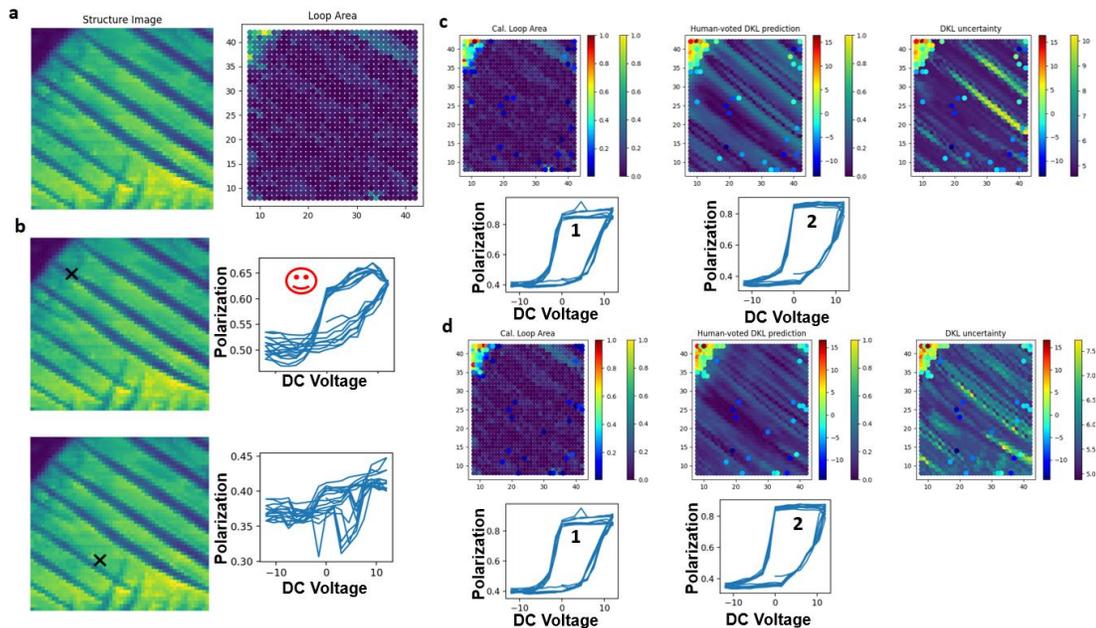

**Fig. 6.** BEPS dataset 2 of PTO thin-film sample as shown in fig (a) with the material structure image design space (left) and the relatively numerically calculated material property such as loop area (right). Fig (b) shows a sample initial comparative human assessment with the preferred location (marked as X) associated with smiley. Figs (c) and (d) are the performance of the px-BO architecture with fully manual and human-AI agent teaming assessments. The left images of these figures are the numerically calculated loop area with the dots representing the same loop area of the explored samples. The middle and the right images of these figures are the predicted mean and variance maps of the utility based on the qualitative assessments (focused on clean spectral with high loop area) with the dots representing the utility of the explored samples.

Finally, we ran the analysis at different $m = 5, 10$ (following Step 2 in Table 1), that is after every 15 and 30 proxy comparisons respectively, to understand the impact on the percentage of human correction throughout the px-BO exploration. **Fig. 7** shows the analysis for both BEPS PTO image space. As expected, we can see for every case, at initial stages, more corrections were required, which swiftly goes down below 10-20% in most cases. This is because, at the early stages, the AI agent is not adequately learned due to small data and gradually learn with more data acquisition. Also, at the initial stages, the chances of discovering unique spectral features (high structural dissimilarity than any of the explored locations) is higher, which the AI-agent is not yet trained for. This indicates the sudden spikes in the correction plots in figs 7a, b and d. Interestingly, in fig 7c, we sudden such numerous spikes which are due to mostly discovering better featured spectral locations, as corrected via human validation. However, we see the average correction range between 16%-25% for all case studies. Thus, the AI agent could accurately assess at least 75% of the time while human correction needed only 25% of the time. In other words, while getting similar optimal results, we could reduce human intervention (human voting) by at least 75% and accelerate discovery via proxy assessment from AI-agent.

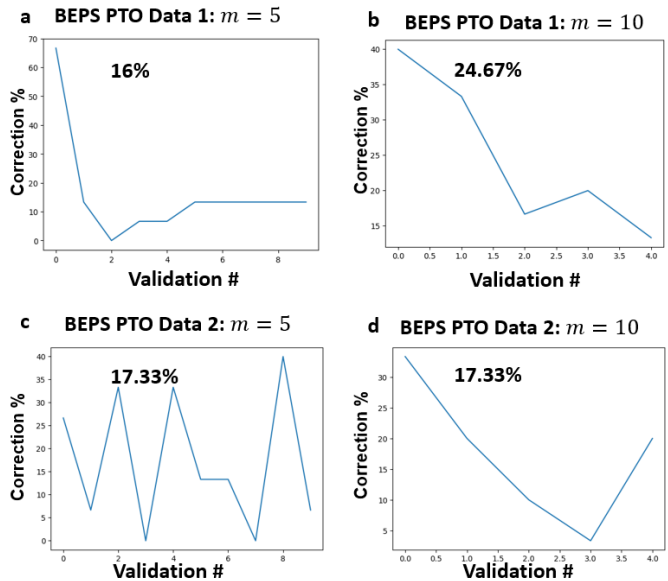

**Fig. 7.** Human correction plots for each autonomous exploration of px-BO over BEPS dataset of PTO thin film sample, at different $m = 5, 10$ (following Step 2, Table 1), i.e. after every 15 and 30 px-BO iterations. Here the X axis is the validation steps during the maximum px-BO iteration of 50, while the Y axis is the percentage of the total non-validated proxy-votes corrected by domain expert. Figs a-b are the analysis done on BEPS dataset referred to in Fig 5. Figs c-d are the analysis done on BEPS dataset referred to in Fig 6.

## 4. CONCLUSION

In summary, we developed a *proxy-modelled Bayesian optimization (px-BO)*, where instead of a traditional numerically calculated objective or scalar metric, we integrated a human-AI collaborative qualitative comparison metric via designing a proxy-system of Bradley-Terry (B-T) model. In this paper for proof of concept, the prototype is presented and demonstrated on simulated images of phases of a ferroelectric material and experimental high-D spectral measurements of a PbTiO3 (PTO) thin film sample. This proxy B-T model serves as transforming the preference between two input locations, based on the quality of phase images and spectral features, into utility-based scores. Then, to minimize human effort, this proxy model also serves as an AI-agent to support human comparative assessment. To ensure quality control of the AI-agent, periodic human validation is also integrated into the design, where we can see the AI-agent continuously learn with BO iterations (more data + human corrections) and get better. Together with human-AI teaming, we see that while humans steer the alignment of autonomous exploration with the domain expert preferred discovery, AI-agents accelerate the exploration towards discovery. Together with the best of both human and AI in AE, we can find better optimal solutions than the traditional purely quantitative function-based AE. As opposed to the proposed design architecture, the standard AE workflows are often non-trivial to numerically formulate (such as loop-area, height, width, storage etc) from experimentally measured complex and noisy spectral structure (which are generally non-smooth than spectral generated from theoretical simulations). As per the future work, the architecture will be connected directly to the microscope for real-time data acquisition and autonomous characterization, finally developing an open-source web app of human-AI teamed AE for usability to broad groups of designers, experimentalists, and material scientists.

**ACKNOWLEDGEMENTS**

This work (A.B) was supported by the University of Tennessee startup funding. The authors acknowledge the use of facilities and instrumentation at the UT Knoxville Institute for Advanced Materials and Manufacturing (IAMM) and the Shull Wollan Center (SWC) supported in part by the National Science Foundation Materials Research Science and Engineering Center program through the UT Knoxville Center for Advanced Materials and Manufacturing (DMR-2309083). This effort (datasets, and ideation of utility proxy and comparative analysis) is supported by the Center for Nanophase Materials Sciences (CNMS), which is a US Department of Energy, Office of Science User Facility at Oak Ridge National Laboratory. This research was sponsored by the INTERSECT Initiative as part of the Laboratory Directed Research and Development Program of Oak Ridge National Laboratory, managed by UT-Battelle, LLC for the US Department of Energy under contract DE-AC05-00OR22725. This work (H.F) was supported by Japan Science and Technology Agency (JST) as part of Adopting Sustainable Partnerships for Innovative Research Ecosystem (ASPIRE),

Grant Number JPMJAP2312, MEXT Program: Data Creation and Utilization Type, Material Research and Development Project (JPMXP1122683430), and MEXT Initiative to Establish Next-generation Novel Integrated Circuits Centers (X-NICS) (JPJ011438).

**Code and Data Availability Statement:**
The analysis reported here along with the code is summarized in Colab Notebook for the purpose of tutorial and application to other data and can be found in https://github.com/arpanbiswas52/papers-code-proxyBO.

**Additional Statement:**
This pre-print is submitted to IDETC, ASME 2026 conference.